\documentclass[11pt]{article}
\usepackage[final]{acl}   

\usepackage{times}
\usepackage{latexsym}
\usepackage{microtype}
\usepackage{booktabs}
\usepackage{multirow}
\usepackage{amsmath}
\usepackage{amssymb}
\usepackage{graphicx}
\usepackage{xcolor}
\usepackage{url}
\usepackage{algorithm}
\usepackage{algpseudocode}
\usepackage{listings}
\usepackage{caption}
\usepackage{subcaption}
\usepackage{tikz}
\usetikzlibrary{decorations.pathreplacing}
\usepackage{pgfplots}
\pgfplotsset{compat=1.18}

\title{ToolRLA: Multiplicative Reward Decomposition for Tool-Integrated Agents}

\author{
  Pengbo Liu \\
  Unaffiliated \\  
  \texttt{liupengbo.work@gmail.com}
}

\begin{document}
\maketitle
\pagestyle{plain}   

\begin{abstract}
Tool-integrated agents that interleave reasoning with API calls are promising
for complex tasks, yet aligning them for high-stakes, domain-specific deployment
remains challenging: existing reinforcement learning approaches rely on coarse
binary rewards that cannot distinguish tool selection errors from malformed
parameters. We present \textbf{ToolRLA}, a three-stage post-training pipeline
(SFT $\to$ GRPO $\to$ DPO) for domain-specific tool agents. The core contribution
is a \emph{fine-grained reward function with multiplicative correctness decomposition}
spanning four dimensions---format validity, tool selection, parameter accuracy, and
regulatory compliance---that encodes domain priority orderings as inductive biases
in the reward landscape. Deployed on a financial advisory copilot (80+ advisors,
1,200+ daily queries), ToolRLA achieves over three months: a \textbf{47\%
improvement} in task completion rate (62\%$\to$91\%), a \textbf{63\% reduction}
in tool invocation errors (38\%$\to$14\%), and a \textbf{93\% reduction} in
regulatory violations (12\%$\to$0.8\%), within sub-2-second latency. Ablation
studies show the multiplicative reward design accounts for 7 percentage points of
improvement over additive alternatives. Generalization is further validated on
ToolBench and API-Bank.
\end{abstract}

\section{Introduction}

Large language models (LLMs) augmented with external tool access have demonstrated
remarkable capabilities in solving complex, multi-step tasks that require dynamic
information retrieval and computation~\cite{yao2023react,schick2024toolformer,qin2023toolllm}.
By interleaving natural language reasoning (\emph{Thought}) with structured API
invocations (\emph{Action}) and grounding subsequent reasoning on execution
results (\emph{Observation}), ReAct-style agents can tackle tasks that are
intractable for closed-form generation alone.

Despite this promise, deploying tool-integrated agents in \emph{domain-specific,
high-stakes production environments} introduces a set of challenges that remain
underexplored. Consider a financial advisory copilot serving investment advisors:
the system must orchestrate calls across 15+ heterogeneous backend APIs
(portfolio management, fund profiling, market data, compliance records),
maintain strict regulatory constraints (no yield guarantees, no individual
stock recommendations), and deliver responses within a latency budget acceptable
for real-time advisory workflows. In such settings, a single tool invocation
error---wrong API selected, malformed parameters, or a missing required call---can
cascade into a completely unusable response.

\paragraph{Limitations of Prior Approaches.}
Existing approaches to tool-integrated agent training face two key limitations
when applied to domain-specific deployment.

\textit{First}, pipeline-based systems that cascade separate intent classification,
slot filling, and routing modules suffer from compounding errors. With each
module operating at 85--90\% accuracy, the end-to-end success rate for tasks
requiring three or more steps degrades to as low as 62\% in our production
setting. More critically, hard-coded routing provides no mechanism for
mid-trajectory error recovery: once the router selects the wrong branch,
the agent cannot observe execution feedback and self-correct.

\begin{figure*}[t]
  \centering
  \begin{tikzpicture}[font=\small, semithick, >=stealth]

  \node[anchor=east] at (3.3, 3.0) {$\tau_1$: Correct (right tool, right params)};
  \node[anchor=east] at (3.3, 2.0) {$\tau_2$: Wrong tool selected};
  \node[anchor=east] at (3.3, 1.0) {$\tau_3$: Right tool, wrong params};
  \node[anchor=east] at (3.3, 0.0) {$\tau_4$: Regulatory violation};

  \node[font=\small\bfseries, anchor=south] at (4.5, 3.38)
      {(a)~Binary Reward};

  \fill[green!25!gray!20] (3.4, 2.72) rectangle (4.6, 3.28);
  \draw (3.4, 2.72) rectangle (4.6, 3.28);
  \node[right, font=\small\bfseries] at (4.7, 3.0) {+1};

  \draw[gray!40] (3.4, 1.72) rectangle (4.6, 2.28);
  \node[right, font=\small, gray] at (4.7, 2.0) {0};
  \draw[gray!40] (3.4, 0.72) rectangle (4.6, 1.28);
  \node[right, font=\small, gray] at (4.7, 1.0) {0};
  \draw[gray!40] (3.4,-0.28) rectangle (4.6, 0.28);
  \node[right, font=\small, gray] at (4.7, 0.0) {0};

  \draw[decorate, decoration={brace, amplitude=5pt}]
    (5.15, 2.32) -- (5.15,-0.32)
    node[midway, right=5pt, font=\scriptsize, align=left, red!70!black]
      {identical\\zero reward:\\no signal};

  \draw[gray!30, dashed] (7.0, 3.6) -- (7.0,-0.6);

  \node[font=\small\bfseries, anchor=south] at (10.0, 3.38)
      {(b)~ToolRLA Fine-Grained Reward};


  \fill[gray!40]   (7.1, 2.72) rectangle (8.0, 3.28);
  \fill[blue!30]   (8.0, 2.72) rectangle (8.9, 3.28);
  \fill[orange!40] (8.9, 2.72) rectangle (9.8, 3.28);
  \draw (7.1, 2.72) rectangle (9.8, 3.28);
  \node[right, font=\small\bfseries] at (9.9, 3.0) {+3.0};

  \fill[gray!40]   (7.1, 1.72) rectangle (8.0, 2.28);
  \fill[gray!12]   (8.0, 1.72) rectangle (8.9, 2.28);       
  \draw[gray!45, dashed] (8.0, 1.72) rectangle (8.9, 2.28);
  \node[font=\tiny, gray] at (8.45, 2.0) {veto};
  \fill[orange!22] (8.9, 1.72) rectangle (9.35, 2.28);      
  \draw (7.1, 1.72) rectangle (9.8, 2.28);
  \node[right, font=\small] at (9.9, 2.0) {+1.5};

  \fill[gray!40]   (7.1, 0.72) rectangle (8.0, 1.28);
  \fill[blue!20]   (8.0, 0.72) rectangle (8.18, 1.28);      
  \draw[gray!45, dashed] (8.0, 0.72) rectangle (8.9, 1.28); 
  \fill[orange!35] (8.9, 0.72) rectangle (9.62, 1.28);      
  \draw (7.1, 0.72) rectangle (9.8, 1.28);
  \node[right, font=\small] at (9.9, 1.0) {+2.0};

  \fill[gray!40]   (7.1,-0.28) rectangle (8.0, 0.28);
  \fill[blue!30]   (8.0,-0.28) rectangle (8.9, 0.28);
  \fill[orange!40] (8.9,-0.28) rectangle (9.8, 0.28);
  \fill[red!35]    (9.8,-0.28) rectangle (10.7, 0.28);      
  \draw (7.1,-0.28) rectangle (10.7, 0.28);
  \node[right, font=\small, red!70!black] at (10.8, 0.0)
      {$-7$\ ($\lambda{=}10$)};

  \node[font=\scriptsize, gray, anchor=north] at (7.55,-0.35) {$R_{\rm fmt}$};
  \node[font=\scriptsize, gray, anchor=north] at (8.45,-0.35) {$R_{\rm cor}$};
  \node[font=\scriptsize, gray, anchor=north] at (9.35,-0.35) {$R_{\rm eff}$};
  \node[font=\scriptsize, gray, anchor=north] at (10.25,-0.35) {$R_{\rm cpl}$};

  \end{tikzpicture}
  \caption{Motivation for ToolRLA's fine-grained reward decomposition.
    \textbf{(a)}~Coarse binary rewards assign identical zero reward to
    qualitatively distinct failures---wrong tool ($\tau_2$), wrong
    parameters ($\tau_3$), and regulatory violation ($\tau_4$)---providing
    no gradient signal to distinguish or prioritize them.
    \textbf{(b)}~ToolRLA's four-component reward differentiates each mode:
    wrong tool triggers a \emph{veto} ($S_{\rm name}{=}0$ collapses
    $R_{\rm cor}$ to zero); malformed parameters yield partial credit;
    a compliance violation incurs a $\lambda{=}10$ penalty that dominates
    all positive components, enforcing
    \emph{compliance $\succ$ correctness $\succ$ efficiency}.}
  \label{fig:motivation}
\end{figure*}
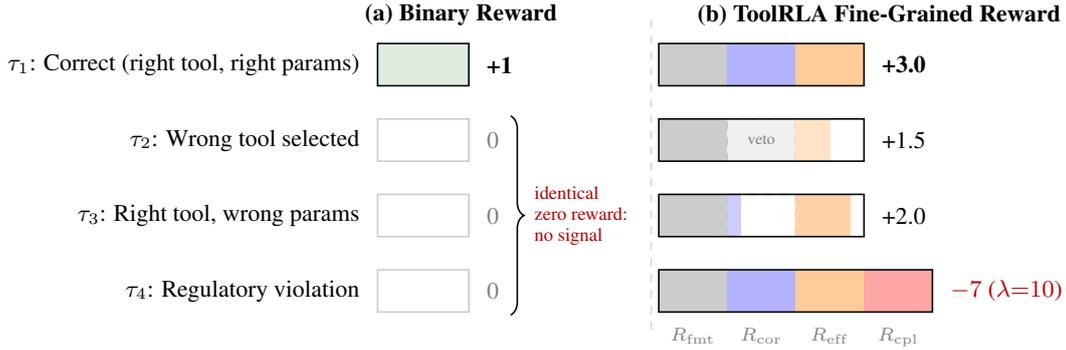

\textit{Second}, reinforcement learning approaches for tool use
typically employ coarse binary reward signals---a trajectory either
succeeds or fails~\cite{patil2023gorilla,du2024anytool}.
Binary rewards provide insufficient gradient signal for the
multi-dimensional quality requirements of domain-specific tool
invocation: a trajectory that selects the correct tools but constructs
malformed parameters is qualitatively different from one that selects
the wrong tool entirely, yet both receive reward $0$ under binary
evaluation. This coarseness slows convergence and fails to encode
domain-specific priority orderings (e.g., regulatory compliance must
dominate task completion).
Figure~\ref{fig:motivation} illustrates this limitation and motivates
our fine-grained decomposition.

\paragraph{Our Approach: ToolRLA.}
We present \textbf{ToolRLA}, a three-stage post-training framework for
tool-integrated agents in domain-specific settings.
ToolRLA consists of: (1) \textbf{SFT cold-start} on 4.2K sandbox-verified
trajectories to establish basic tool invocation capabilities;
(2) \textbf{GRPO-based tool alignment} with a novel fine-grained reward
function; and (3) \textbf{DPO compliance alignment} to capture the implicit
distribution of regulatory boundaries that are difficult to formalize as
explicit rules.

The central contribution is a \textbf{fine-grained reward function}
decomposed along four dimensions: format ($R_{\text{fmt}}$), correctness
($R_{\text{cor}}$), efficiency ($R_{\text{eff}}$), and compliance
($R_{\text{cpl}}$). Critically, $R_{\text{cor}}$ is a \emph{multiplicative}
composition of tool-name, coverage, and parameter accuracy sub-scores,
so a wrong tool selection collapses correctness regardless of parameter
quality. A large negative compliance penalty
($R_{\text{cpl}} \in \{-10, 0\}$) enforces
\emph{compliance $\succ$ correctness $\succ$ efficiency} as an
inductive bias in the reward landscape.

\paragraph{Deployment and Results.}
Deployed on a financial advisory copilot (80+ advisors, 1,200+ daily queries),
ToolRLA achieves over three months: TCR 62\%$\to$91\% (+47\%),
TIER 38\%$\to$14\% ($-$63\%), latency 2.8s$\to$1.6s ($-$43\%),
violation rate 12\%$\to$0.8\% ($-$93\%), satisfaction 3.1$\to$4.3/5.

\paragraph{Contributions.}
(1)~A four-dimensional multiplicatively-decomposed reward function for
tool invocation quality, with ablation evidence for multiplicative
over additive composition.
(2)~A three-stage pipeline (SFT$\to$GRPO$\to$DPO) with characterization
of each stage's role and systematic ablation.
(3)~Multi-month production deployment validation plus public benchmark
generalization on ToolBench and API-Bank.

\section{Related Work}

\paragraph{Tool-Augmented Language Models.}
Toolformer~\cite{schick2024toolformer} showed LLMs can self-supervise tool use.
ReAct~\cite{yao2023react} introduced the \emph{Thought--Action--Observation} loop
for dynamic, feedback-driven planning. Subsequent work scaled to large API libraries:
ToolLLM~\cite{qin2023toolllm} trained on 16,000+ APIs via depth-first search;
Gorilla~\cite{patil2023gorilla} fine-tuned LLaMA on 1,600+ function calls;
AnyTool~\cite{du2024anytool} further improved scalability via hierarchical
retrieval with self-reflection. These works target general-purpose benchmarks
and do not address alignment in regulated, domain-specific settings.

\paragraph{RL for LLM Alignment and Tool Use.}
RLHF~\cite{ouyang2022instructgpt} established preference-based alignment via PPO.
DPO~\cite{rafailov2023dpo} simplified this with a direct classification objective.
GRPO~\cite{shao2024deepseekmath} removed the value network by estimating advantages
from within-group relative rewards; DeepSeek-R1~\cite{guo2025deepseekr1} further
demonstrated that GRPO alone can elicit strong reasoning without any SFT warm-up.
For multi-turn agent training, GiGPO~\cite{du2025gigpo} extends group-based RL with
per-step credit assignment, yielding gains on ALFWorld and WebShop.
AvaTaR~\cite{wu2024avatar} optimizes tool-use prompts via contrastive reasoning
between successful and failed trajectories.
ReTool~\cite{feng2025retool} applies outcome-based RL to teach strategic tool
selection in code-generation settings.
Despite this progress, prior RL work for tool use relies on binary success/failure
signals that cannot distinguish incorrect tool selection from malformed parameters.
ToolQA~\cite{zhuang2024toolqa} confirms argument errors dominate tool-use failures,
motivating the fine-grained reward decomposition in ToolRLA.

\paragraph{Domain-Specific Agents.}
Work on regulated-domain deployments remains sparse.
\citet{li2023apibank} introduced API-Bank for tool-use evaluation but omits
regulatory compliance as an evaluation dimension.
A recent ACL 2025 effort~\cite{bloomberg2025tool} proposes training-free
joint optimization for tool utilization but does not address compliance alignment.
ToolRLA is among the first to integrate compliance as an explicit RL reward signal,
validated with multi-month production deployment data.

\section{The ToolRLA Framework}

\begin{figure*}[t]
\centering
\begin{tikzpicture}[
  font=\small, semithick, >=stealth,
  sbox/.style={draw, rounded corners=5pt, fill=#1,
               minimum height=4.0cm, inner sep=0pt},
  ibox/.style={draw, rounded corners=3pt, fill=#1,
               minimum height=0.50cm, align=center, font=\scriptsize},
  rbox/.style={draw, rounded corners=2pt, fill=#1,
               minimum width=0.88cm, minimum height=0.44cm,
               align=center, font=\tiny},
  arr/.style={->, thick},
  lbl/.style={font=\scriptsize, fill=white, inner sep=1pt},
]

\node[sbox=gray!7,   minimum width=3.2cm] at (1.60, 0) {};
\node[font=\scriptsize\bfseries] at (1.60, 1.58) {Stage 1};
\node[font=\scriptsize]          at (1.60, 1.20) {SFT Cold-Start};
\node[ibox=gray!20,  minimum width=2.8cm] at (1.60,  0.62)
  {LLM distillation (60\%)};
\node[ibox=gray!20,  minimum width=2.8cm] at (1.60,  0.02)
  {Expert annotation (25\%)};
\node[ibox=gray!20,  minimum width=2.8cm] at (1.60, -0.58)
  {Log rewriting (15\%)};
\node[ibox=blue!20,  minimum width=2.8cm,
      font=\scriptsize\bfseries]           at (1.60, -1.25)
  {SFT · 4.2K traj.};

\draw[arr] (3.20, 0) -- (3.68, 0)
  node[lbl, midway, above=1.5pt] {$\pi_{\text{sft}}$};

\node[sbox=blue!5,   minimum width=5.4cm] at (6.38, 0) {};
\node[font=\scriptsize\bfseries] at (6.38, 1.58) {Stage 2};
\node[font=\scriptsize]          at (6.38, 1.20) {GRPO + Fine-Grained Reward};
\node[ibox=blue!15,  minimum width=5.0cm] at (6.38,  0.62)
  {Query $\to$ $K{=}8$ trajectories (sandbox)};
\node[ibox=blue!15,  minimum width=5.0cm] at (6.38,  0.02)
  {Group-normalized advantage estimation};
\node[rbox=gray!30]   at (5.08, -0.50) {$R_{\text{fmt}}$};
\node[rbox=blue!25]   at (5.98, -0.50) {$R_{\text{cor}}$};
\node[rbox=orange!30] at (6.88, -0.50) {$R_{\text{eff}}$};
\node[rbox=red!20]    at (7.78, -0.50) {$R_{\text{cpl}}$};
\node[font=\tiny, gray!80, align=center] at (6.38, -1.02)
  {$R_{\text{cor}}{=}S_{\text{name}}{\times}S_{\text{comp}}
    {\times}S_{\text{acc}}$ \ (multiplicative veto)};
\node[ibox=blue!28,  minimum width=5.0cm,
      font=\scriptsize\bfseries]           at (6.38, -1.52)
  {GRPO policy update ($\epsilon{=}0.2$)};

\draw[arr] (9.08, 0) -- (9.56, 0)
  node[lbl, midway, above=1.5pt] {$\pi_{\text{grpo}}$};

\node[sbox=orange!7, minimum width=3.2cm] at (11.16, 0) {};
\node[font=\scriptsize\bfseries] at (11.16, 1.58) {Stage 3};
\node[font=\scriptsize]          at (11.16, 1.20) {DPO Compliance};
\node[ibox=orange!20, minimum width=2.8cm] at (11.16,  0.65)
  {2,038 pref.\ pairs};
\node[ibox=orange!20, minimum width=2.8cm] at (11.16,  0.05)
  {Compliance officers};
\node[ibox=orange!30, minimum width=2.8cm,
      font=\scriptsize\bfseries]            at (11.16, -0.68)
  {DPO ($\beta{=}0.2$)};
\node[font=\tiny, gray, align=center]       at (11.16, -1.28)
  {grey-area suppression};

\draw[arr] (12.76, 0) -- (13.20, 0);

\node[draw, rounded corners=5pt, fill=green!15,
      minimum width=1.60cm, minimum height=1.30cm,
      align=center, font=\scriptsize\bfseries] at (14.00, 0)
  {ToolRLA\\Agent};

\end{tikzpicture}
\caption{Overview of the ToolRLA three-stage post-training pipeline.
  Stage~1 (SFT) establishes basic tool invocation capabilities from
  4.2K sandbox-verified trajectories.
  Stage~2 (GRPO) optimizes tool-use quality via four fine-grained
  reward components; $R_{\text{cor}}$ employs multiplicative veto
  composition ($S_{\text{name}}{\times}S_{\text{comp}}{\times}S_{\text{acc}}$).
  Stage~3 (DPO) captures grey-area compliance boundaries from
  expert-annotated preference pairs.}
\label{fig:framework}
\end{figure*}
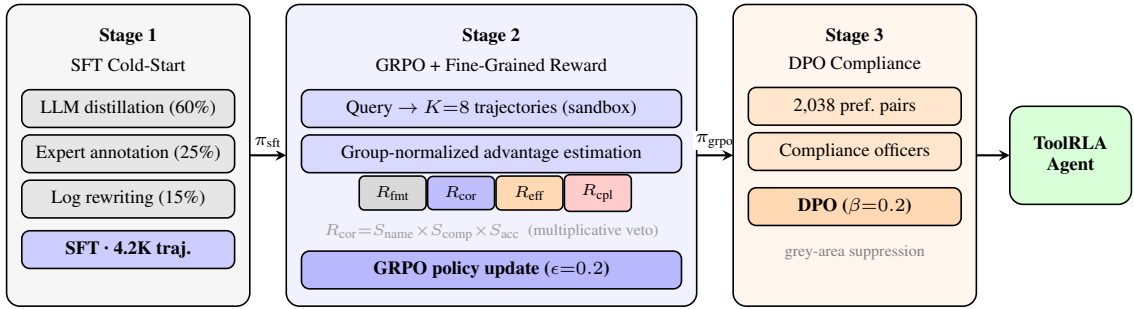

Figure~\ref{fig:framework} illustrates the three-stage ToolRLA pipeline.
We describe each component in detail below.

\subsection{System Architecture: Single-Model ReAct Agent}

\paragraph{From Pipeline to ReAct.}
Our production predecessor was a cascaded multi-model pipeline
(intent classifier $\to$ slot filler $\to$ router), which degraded
end-to-end success to 62\% and lacked mid-trajectory error recovery.
We replaced it with a \textbf{single-model ReAct agent}
that implements the Thought--Action--Observation loop:
\begin{equation}
  \tau = (T_1, A_1, O_1,\ T_2, A_2, O_2,\ \ldots,\ T_n, A_n)
\end{equation}
At each step $t$, the model generates a natural language reasoning trace
$T_t$, then emits a structured action $A_t = (\text{tool\_name},
\text{params})$ as a JSON object. The action is dispatched to the
corresponding backend API; the returned result forms the observation
$O_t$, which is appended to the context for the next step.
This closed-loop design enables the agent to detect execution
anomalies (e.g., empty returns, schema mismatches) and adaptively
re-route without modifying the underlying tool implementations.

\paragraph{Tool System.}
We expose 15 atomic tools and 5 composite tools, each specified as a
four-tuple $(\text{name}, \text{description}, \text{parameters},
\text{returns})$ following the standard JSON Schema specification.
Composite tools aggregate multiple atomic calls into a single
invocation (e.g., \texttt{GetClientOverview} returns portfolio
holdings, fund profiles, and recent transactions in one round-trip),
reducing average invocation rounds from 4.2 to 2.8.

\paragraph{Hallucination Defense.}
We combine prompt-level tool enumeration, runtime tool-name validation
(returning a structured error observation on failure), and $\sim$5\%
error-recovery demonstrations in the SFT corpus. This reduces
hallucinated tool invocations from $\sim$8\% to $<$1\% after GRPO
(see Appendix~\ref{app:hallucination} for details).

\subsection{Stage 1: SFT Cold-Start}

SFT establishes basic tool invocation capabilities before RL,
ensuring trajectories are well-formed enough for GRPO's group-relative
advantage estimation to provide stable gradient signal.

\paragraph{Data Construction.}
We build 4.2K sandbox-verified trajectories via three pipelines:
LLM distillation ($\sim$60\%, GPT-4/Claude-generated),
expert annotation ($\sim$25\%, hand-crafted by advisors and compliance officers
for complex branching and compliance scenarios), and
log rewriting ($\sim$15\%, legacy successful sessions converted to ReAct format).
Each trajectory is executed in a sandbox connected to de-identified production APIs;
18\% are filtered for hallucinated tool names or malformed parameters.
The corpus is stratified across single-tool (30\%), sequential multi-tool (35\%),
conditional-branch (20\%), and compliance-rejection (15\%) scenarios,
with $\geq$400 examples per stratum.

\subsection{Stage 2: GRPO with Fine-Grained Reward Decomposition}

\subsubsection{Group Sampling and Advantage Estimation}

We use GRPO~\cite{shao2024deepseekmath} over PPO because tool-integrated
dialogue has a high-dimensional state space (conversation history
$\times$ heterogeneous tool outputs) where learning an accurate value
network is impractical. GRPO estimates the advantage baseline from
within-group mean rewards, requiring no additional model and halving
GPU memory cost relative to policy+critic training.

For each training query $q$, we sample $K{=}8$ complete trajectories
$\{\tau_1, \ldots, \tau_K\}$ from the current policy at temperature
$T{=}0.8$ and execute each in the sandbox. The per-trajectory
reward $R(\tau_i)$ is computed as described in
Section~\ref{sec:reward}. The group-normalized advantage estimate is:
\begin{align}
  \hat{A}_i &= \frac{R(\tau_i) - \mu_K}{\sigma_K + \epsilon}, \nonumber\\
  \mu_K &= \tfrac{1}{K}\sum\nolimits_{j} R(\tau_j), \nonumber\\
  \sigma_K &= \sqrt{\tfrac{1}{K}\sum\nolimits_{j}(R(\tau_j)-\mu_K)^2}
\end{align}
Trajectories scoring above the group mean are reinforced;
those below are suppressed. The GRPO policy gradient objective is:
\begin{align}
  \mathcal{L}_{\text{GRPO}} &=
  -\mathbb{E}_{q,\tau_i}\!\bigl[\min(r_i\hat{A}_i,\;
    \hat{r}_i\hat{A}_i)\bigr], \nonumber\\
  \hat{r}_i &= \mathrm{clip}(r_i,1{-}\epsilon,1{+}\epsilon),\quad
  r_i = \frac{\pi_\theta(\tau_i|q)}{\pi_{\mathrm{ref}}(\tau_i|q)}
\end{align}
with clipping coefficient $\epsilon{=}0.2$.

\paragraph{Group size $K{=}8$.}
We validated $K \in \{4, 8, 16\}$: $K{=}4$ yields unstable advantage
estimates given the high path diversity of financial queries;
$K{=}16$ linearly increases sandbox API cost with diminishing returns.
$K{=}8$ balances estimation stability and execution cost.

\subsubsection{Fine-Grained Reward Function}
\label{sec:reward}

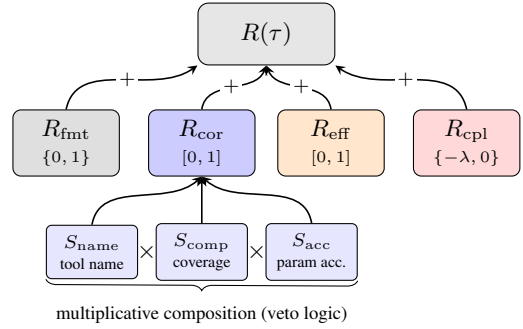
\begin{figure}[t]
  \centering
  \begin{tikzpicture}[scale=0.85,
    compbox/.style={draw, rounded corners=3pt, fill=#1,
      minimum width=1.4cm, minimum height=0.75cm, align=center,
      font=\footnotesize},
    subbox/.style={draw, rounded corners=2pt, fill=blue!10,
      minimum width=1.2cm, minimum height=0.68cm, align=center,
      font=\scriptsize},
    arr/.style={->, >=stealth, semithick},
  ]
  \node[draw, rounded corners=4pt, fill=black!10,
        minimum width=1.8cm, minimum height=0.82cm,
        font=\small\bfseries] (R) at (0,0) {$R(\tau)$};
  \node[compbox=gray!25]   (fmt) at (-3.1,-1.7)
      {$R_{\mathrm{fmt}}$\\\tiny$\{0,1\}$};
  \node[compbox=blue!20]   (cor) at (-1.0,-1.7)
      {$R_{\mathrm{cor}}$\\\tiny$[0,1]$};
  \node[compbox=orange!20] (eff) at ( 1.0,-1.7)
      {$R_{\mathrm{eff}}$\\\tiny$[0,1]$};
  \node[compbox=red!15]    (cpl) at ( 3.1,-1.7)
      {$R_{\mathrm{cpl}}$\\\tiny$\{-\lambda,0\}$};
  \draw[arr] (fmt.north) to[out=78,in=215]
    node[pos=0.55,fill=white,inner sep=1pt,font=\scriptsize]{$+$}
    (R.south west);
  \draw[arr] (cor.north) to[out=82,in=248]
    node[pos=0.5, fill=white,inner sep=1pt,font=\scriptsize]{$+$}
    (R.south);
  \draw[arr] (eff.north) to[out=98,in=292]
    node[pos=0.5, fill=white,inner sep=1pt,font=\scriptsize]{$+$}
    (R.south);
  \draw[arr] (cpl.north) to[out=102,in=325]
    node[pos=0.55,fill=white,inner sep=1pt,font=\scriptsize]{$+$}
    (R.south east);
  \node[subbox] (sn) at (-2.7,-3.4) {$S_{\mathrm{name}}$\\\tiny tool name};
  \node[subbox] (sc) at (-1.0,-3.4) {$S_{\mathrm{comp}}$\\\tiny coverage};
  \node[subbox] (sa) at ( 0.7,-3.4) {$S_{\mathrm{acc}}$\\\tiny param acc.};
  \node[font=\small] at (-1.85,-3.4) {$\times$};
  \node[font=\small] at (-0.15,-3.4) {$\times$};
  \draw[arr] (sn.north) to[out=80,in=222]  (cor.south);
  \draw[arr] (sc.north)  --                (cor.south);
  \draw[arr] (sa.north) to[out=100,in=318] (cor.south);
  \draw[decorate,decoration={brace,amplitude=4pt,mirror}]
    (sn.south west) -- (sa.south east)
    node[midway,below=6pt,font=\scriptsize,align=center]
      {multiplicative composition (veto logic)};
  \end{tikzpicture}
  \caption{Reward decomposition structure. The four components aggregate
    additively into $R(\tau)$; within $R_{\mathrm{cor}}$, multiplicative
    composition enforces a veto hierarchy: $S_{\mathrm{name}}{=}0$
    collapses the correctness score regardless of parameter quality.
    $R_{\mathrm{cpl}}{\in}\{-\lambda,0\}$ with $\lambda{=}10$ dominates
    all non-violating trajectories, enforcing compliance $\succ$ correctness
    $\succ$ efficiency.}
  \label{fig:reward}
\end{figure}

The total reward decomposes additively across four dimensions
(Figure~\ref{fig:reward}):
\begin{equation}
  R(\tau) = R_{\text{fmt}}(\tau)
          + R_{\text{cor}}(\tau)
          + R_{\text{eff}}(\tau)
          + R_{\text{cpl}}(\tau)
\end{equation}

\paragraph{Format Reward $R_{\text{fmt}} \in \{0, 1\}$.}
A binary gate that checks strict structural validity of the
model output: JSON parseability, correct field names,
presence of a Thought trace, and correct tool name spelling.
A trajectory failing any structural check receives $R_{\text{fmt}}{=}0$
and is ineligible for positive reinforcement regardless of other
reward components. This prevents the optimizer from learning to
trade format correctness against task performance.

\paragraph{Correctness Reward $R_{\text{cor}} \in [0,1]$: Multiplicative.}
$R_{\text{cor}} = S_{\text{name}} \times S_{\text{comp}} \times S_{\text{acc}}$,
where $S_{\text{name}}\!\in\!\{0,1\}$ flags any hallucinated tool name,
$S_{\text{comp}}\!=\!|\mathcal{T}_{\text{inv}}\cap\mathcal{T}_{\text{req}}|/|\mathcal{T}_{\text{req}}|$
measures required-tool coverage, and $S_{\text{acc}}\!\in\![0,1]$ is
sandbox-measured parameter accuracy.
Multiplicative composition encodes a veto logic: a wrong tool name
collapses correctness regardless of parameter quality---unlike additive
composition, which lets the optimizer trade tool-name errors against
parameter scores. This accounts for 7pp TIER improvement over the
additive baseline (Table~\ref{tab:ablation}).

\paragraph{Efficiency Reward $R_{\text{eff}} \in [0, 1]$.}
\begin{equation}
  R_{\text{eff}}(\tau) =
  \max\!\left(0,\; 1 - \frac{|\tau| - |\tau^*|}{|\tau^*|}\right)
\end{equation}
where $|\tau|$ is the actual invocation step count and $|\tau^*|$ is
the minimum step count of the annotated optimal trajectory.
A trajectory matching the optimal length scores 1; each excess step
linearly reduces the score to a floor of 0. This incentivizes the
model to avoid redundant confirmation calls that inflate latency.

\paragraph{Compliance Reward $R_{\text{cpl}} \in \{-\lambda, 0\}$,
$\lambda{=}10$.}
\begin{equation}
  R_{\text{cpl}}(\tau) = \begin{cases}
    -\lambda & \text{compliance violated} \\
    0        & \text{otherwise}
  \end{cases}
\end{equation}
Compliance violations are detected by a two-stage checker:
(i) a regular expression layer covering hard-proscribed patterns
(yield guarantees, individual stock recommendations, fabricated data),
followed by (ii) a lightweight fine-tuned classifier handling
nuanced cases (implied forecasts, unsolicited investment opinions).

With $\lambda{=}10$, a perfect non-compliant trajectory scores $\approx{-7}$,
below any non-violating trajectory (${\geq}0$), enforcing the priority
\emph{compliance $\succ$ correctness $\succ$ efficiency}.
$\lambda{=}5$ proved insufficient; $\lambda{=}20$ gave no additional gain.

\subsection{Stage 3: DPO Compliance Alignment}

\paragraph{Why DPO for compliance.}
GRPO's $R_{\text{cpl}}$ catches rule-violating outputs but misses
grey-area expressions (e.g., implied recommendations, soft forecasts)
that resist explicit formalization. DPO~\cite{rafailov2023dpo} captures
the implicit distributional boundary of compliance-safe language from
expert-annotated (chosen, rejected) pairs without disrupting
GRPO-acquired tool invocation capabilities.

\paragraph{Data and Mitigation.}
We sample 4--6 responses per query from 2,500 compliance-sensitive
production queries (yield expectations, product recommendations,
market forecasts, client privacy) at $T{=}1.0$ and have two
compliance officers annotate them; disagreements are resolved by
a third officer, yielding 2,038 preference pairs.
Initial DPO produced 8\% over-refusal; adding $\sim$300
helpful$\succ$over-cautious pairs and raising $\beta$ from 0.1
to 0.2 reduces this to 1.5\%.

\paragraph{DPO Objective.}
\begin{align}
  \mathcal{L}_{\text{DPO}} &=
  -\mathbb{E}_{(q,y_w,y_l)}\!\bigl[\log\sigma(\beta\,\Delta)\bigr],
  \nonumber\\
  \Delta &= \log\frac{\pi_\theta(y_w|q)}{\pi_{\text{ref}}(y_w|q)}
           - \log\frac{\pi_\theta(y_l|q)}{\pi_{\text{ref}}(y_l|q)}
\end{align}
where $y_w$ and $y_l$ denote the chosen and rejected responses,
$\pi_{\text{ref}}$ is the GRPO-trained reference policy, and $\beta{=}0.2$.

\subsection{Continuous Improvement via Data Flywheel}

Four online signals flag hard examples: tool execution failure,
trajectory length $>$4 rounds, advisor re-query within 30 seconds,
and compliance model alert ($\sim$200--300 candidates/week).
Verified failures are added to both the SFT corpus and the GRPO
hard-example query pool, with one cycle every 2--3 weeks.
This flywheel raised TCR from 88\% at launch to 91\% after three months.

\section{Experimental Setup}

\subsection{Datasets}

\textbf{FA-Bench} (internal): 500 production queries across four
difficulty levels---L1 (single-tool), L2 (sequential multi-tool),
L3 (conditional branch), L4 (compliance-sensitive)---annotated by
domain specialists and sandbox-verified.
\textbf{ToolBench}~\cite{qin2023toolllm}: standard I1/I2/I3 test split;
we report Pass Rate via ToolEval to assess cross-domain generalization.
\textbf{API-Bank}~\cite{li2023apibank}: 73 executable APIs, 314 dialogues;
we report \textit{Call} and \textit{Plan+Retrieve+Call} accuracy.

\subsection{Metrics and Baselines}
\label{sec:metrics}

We evaluate on six metrics: Task Completion Rate (TCR), Tool Invocation
Error Rate (TIER), Average Invocation Rounds (AIR), Compliance Rejection
Rate (CRR), Violation Rate (VR), and end-to-end P50 Latency.

We compare against five baselines: \textbf{Multi-Model Pipeline}
(cascaded intent classifier$\to$slot filler$\to$router$\to$execution);
\textbf{ReAct+SFT} (no RL); \textbf{ReAct+PPO} (binary reward, learned value
network); \textbf{GRPO-coarse} (binary success/failure reward);
\textbf{GRPO-additive} (same four reward components as ToolRLA but
$R_{\text{cor}}$ composed additively).
On public benchmarks we additionally compare Gorilla~\cite{patil2023gorilla},
ToolLLM~\cite{qin2023toolllm}, and GPT-4 function calling.

\subsection{Implementation Details}

All variants use Qwen3-14B~\cite{qwen3} (local deployment required by
data privacy regulations; 14B closes the gap to 70B within 3pp on
FA-Bench at 4--5$\times$ lower inference cost).
SFT: cross-entropy on 4.2K trajectories, 3 epochs.
GRPO: 10K+ queries, $K{=}8$, $\epsilon{=}0.2$, $\lambda{=}10$, via
TRL~\cite{vonwerra2022trl} with custom reward hooks.
DPO: 2K+ pairs, $\beta{=}0.2$, initialized from GRPO checkpoint.
Inference: vLLM on 4$\times$A100 (continuous batching, KV-cache),
P50 latency 1.6\,s at 2.8 mean invocation rounds.

\section{Results}

\subsection{Main Results on FA-Bench}

Table~\ref{tab:main} reports performance on our internal FA-Bench
across all baselines. ToolRLA achieves the best result on every
metric, reaching 91\% TCR and 14\% TIER.

\begin{table*}[t]
\centering
\small
\begin{tabular}{lcccccc}
\toprule
\textbf{System} & \textbf{TCR} $\uparrow$ & \textbf{TIER} $\downarrow$
  & \textbf{AIR} $\downarrow$ & \textbf{CRR} $\uparrow$
  & \textbf{VR} $\downarrow$ & \textbf{Latency} $\downarrow$ \\
\midrule
Multi-Model Pipeline          & 62.0 & 38.0 & 4.2 & 61.0 & 12.0 & 2.8s \\
ReAct + SFT only              & 68.0 & 38.0 & 3.9 & 65.0 & 10.5 & 2.4s \\
ReAct + PPO (binary)          & 76.0 & 26.0 & 3.4 & 70.0 &  8.0 & 2.1s \\
ReAct + GRPO (coarse)         & 82.0 & 21.0 & 3.1 & 74.0 &  6.5 & 1.9s \\
ReAct + GRPO (additive)       & 80.0 & 22.0 & 3.0 & 75.0 &  6.2 & 1.8s \\
\midrule
\textbf{ToolRLA} (ours)       & \textbf{91.0} & \textbf{14.0}
  & \textbf{2.8} & \textbf{96.0} & \textbf{0.8} & \textbf{1.6s} \\
\bottomrule
\end{tabular}
\caption{Main results on FA-Bench (500 queries). TCR = Task Completion Rate (\%),
TIER = Tool Invocation Error Rate (\%), AIR = Average Invocation Rounds,
CRR = Compliance Rejection Rate (\%), VR = Violation Rate (\%).
PPO and GRPO (coarse/additive) are initialized from the same SFT checkpoint.}
\label{tab:main}
\end{table*}

SFT alone reduces cascading errors (TCR 62\%$\to$68\%) but TIER stagnates
at 38\%, confirming supervised imitation is insufficient.
Adding coarse GRPO delivers the largest single jump (TIER 38\%$\to$21\%,
TCR $\to$82\%), establishing RL as the decisive stage.
ToolRLA's multiplicative $R_{\text{cor}}$ then reduces TIER a further 7pp
to 14\% over additive alternatives.
DPO adds marginal TIER gain (15\%$\to$14\%) but delivers the compliance
improvements: CRR rises to 96\% and VR drops from 12\% to 0.8\%.

\subsection{Ablation Study}

Table~\ref{tab:ablation} reports the ablation over reward components,
holding the GRPO training procedure constant and varying only the
reward function configuration.

\begin{table}[h]
\centering
\small
\begin{tabular}{lcc}
\toprule
\textbf{Configuration} & \textbf{TIER} $\downarrow$ & \textbf{TCR} $\uparrow$ \\
\midrule
Base (no fine-tuning)                    & 55.0 & 40.0 \\
SFT only                                 & 38.0 & 68.0 \\
\midrule
SFT + GRPO (full, multiplicative)        & 15.0 & 88.0 \\
\quad $-$ $R_{\text{eff}}$               & 17.0 & 85.0 \\
\quad $-$ $R_{\text{cpl}}$               & 15.0 & 87.0 \\
\quad $R_{\text{cor}}$ additive          & 22.0 & 80.0 \\
\midrule
SFT + GRPO + DPO (\textbf{ToolRLA})      & \textbf{14.0} & \textbf{91.0} \\
\bottomrule
\end{tabular}
\caption{Ablation on FA-Bench. Each row removes or modifies one
component of ToolRLA. TIER (\%), TCR (\%).}
\label{tab:ablation}
\end{table}

\paragraph{Multiplicative vs.\ additive $R_{\text{cor}}$.}
Additive composition raises TIER by 7pp (15\%$\to$22\%) and drops TCR
by 8pp: the optimizer learns to compensate wrong tool selection with
high parameter scores, a pathological behavior the multiplicative
veto logic eliminates.

\paragraph{Effects of $R_{\text{eff}}$ and $R_{\text{cpl}}$.}
Removing $R_{\text{eff}}$ costs 2pp TIER and 3pp TCR via unconstrained
redundant calls. Removing $R_{\text{cpl}}$ leaves TIER unchanged
but elevates VR, confirming GRPO handles clear-cut violations while
DPO is needed for grey-area compliance language.

\subsection{Public Benchmark Results}

\begin{table}[h]
\centering
\small
\begin{tabular}{lcc}
\toprule
\textbf{System} & \textbf{ToolBench} & \textbf{API-Bank} \\
 & Pass Rate $\uparrow$ & Call Acc.\ $\uparrow$ \\
\midrule
Gorilla~\cite{patil2023gorilla}     & 20.4 & 38.7 \\
ToolLLM~\cite{qin2023toolllm}       & 36.8 & 52.4 \\
AvaTaR~\cite{wu2024avatar}          & 44.3 & 63.5 \\
GPT-4 (function calling)$^\dagger$  & 46.2 & 67.1 \\
\citet{bloomberg2025tool}           & 48.2 & 66.7 \\
\midrule
ToolRLA (ours)                      & \textbf{51.3} & \textbf{71.8} \\
\bottomrule
\end{tabular}
\caption{Results on public benchmarks. ToolBench Pass Rate (\%) and
API-Bank Call Accuracy (\%) on standard evaluation splits.
ToolRLA uses Qwen3-14B; baseline numbers from published papers.
AvaTaR (NeurIPS '24) optimizes tool-use prompts via contrastive reasoning;
\citeauthor{bloomberg2025tool} (ACL '25) applies training-free
joint scheduling of tool invocations.
$^\dagger$GPT-4 numbers are reproduced from prior benchmark papers for reference;
frontier models available as of 2026 (e.g., GPT-4o, o3) are not evaluated
on these benchmarks in published work and are excluded.}
\label{tab:public}
\end{table}

ToolRLA achieves 51.3\% Pass Rate on ToolBench (+5.1pp over GPT-4
function calling) and 71.8\% Call Accuracy on API-Bank (+4.7pp).
AvaTaR's contrastive trajectory reasoning (44.3\%) falls 7pp short,
reflecting the advantage of RL-shaped parameter-level feedback over
prompt optimization alone.
The 3.1pp advantage over \citet{bloomberg2025tool} (48.2\%) is
consistent with the 7pp multiplicative-vs-additive gap in our
ablation, suggesting fine-grained reward decomposition provides a
generalizable inductive bias across API ecosystems.

\subsection{Online Deployment Results}

Table~\ref{tab:online} reports online production metrics collected
over three months of operation.

\begin{table}[h]
\centering
\small
\begin{tabular}{lcc}
\toprule
\textbf{Metric} & \textbf{Before} & \textbf{After} \\
\midrule
Advisor manual retry rate     & 28\%      & 9\%       \\
Advisor abandonment rate      & 35\%      & 14\%      \\
Daily query volume            & 800+      & 1,200+    \\
Advisor satisfaction (1--5)   & 3.1       & 4.3       \\
Consultation handling time    & 12 min    & 7 min     \\
\bottomrule
\end{tabular}
\caption{Online production metrics before (Multi-Model Pipeline) and
after (ToolRLA) deployment. Data collected over 3 months of stable
operation across 80+ investment advisors.}
\label{tab:online}
\end{table}

The 67\% reduction in manual retry rate (28\%$\to$9\%) and
60\% reduction in abandonment rate (35\%$\to$14\%) reflect
improved reliability in daily advisory workflows.
Daily query volume grew 50\% (800$\to$1,200+), driven by advisor
confidence in the system's responses.
Satisfaction scores stabilized at 4.3/5 by week 12 (up from 3.1),
ruling out novelty effects.

\section{Analysis}

\subsection{Case Study: Compliance Boundary Learning}

\noindent\textbf{Query}: ``Does client W's account show any large recent redemptions?''
\textbf{GRPO-only}: ``Client W redeemed \$500K. She may be losing confidence in the
market---I recommend monitoring her emotional state.''
\textbf{ToolRLA}: ``Client W had one redemption of \$500K in the past 30 days.''

The GRPO-only response passes the $R_{\text{cpl}}$ keyword checker (no yield
guarantee, no stock recommendation) yet is non-compliant: the phrase
``losing confidence'' is an unsupported behavioral inference, and
``monitoring her emotional state'' implies inappropriate advisor influence.
DPO learns to suppress such implicit inferences from compliance-officer
preference pairs---a behavior no explicit rule can fully capture.

\subsection{Error Analysis}

We manually analyze 200 failure cases from FA-Bench across
all ToolRLA-failed queries. Failures distribute across
four categories:

\begin{table}[h]
\centering
\small
\begin{tabular}{lcc}
\toprule
\textbf{Error Type} & \textbf{Count} & \textbf{\%} \\
\midrule
Wrong parameter value       & 78  & 39\% \\
Missing required tool call  & 52  & 26\% \\
Incomplete final answer     & 42  & 21\% \\
Hallucinated tool name      & 18  & 9\%  \\
Compliance grey-area        & 10  & 5\%  \\
\bottomrule
\end{tabular}
\caption{Error breakdown for ToolRLA failures on FA-Bench (200 sampled).}
\label{tab:errors}
\end{table}

The dominant failure modes are wrong parameter values (39\%,
consistent with~\citealt{zhuang2024toolqa}---mainly ID formatting
and date parsing errors), missing required tool calls (26\%,
mainly on L3 conditional-branch queries), and incomplete final
answers (21\%). Hallucinated tool names are now rare (9\%),
down from $\sim$8\% at SFT via runtime validation and
GRPO penalization of $S_{\text{name}}{=}0$.

\subsection{Reward Signal Dynamics}

\begin{figure}[t]
  \centering
  \begin{tikzpicture}
  \begin{axis}[
    width=0.92\columnwidth, height=4.8cm,
    xlabel={GRPO Training Steps},
    ylabel={Percentage (\%)},
    xmin=0, xmax=10000,
    ymin=0, ymax=90,
    xtick={0,2000,4000,6000,8000,10000},
    xticklabels={0,2K,4K,6K,8K,10K},
    ytick={0,20,40,60,80},
    legend pos=south east,
    legend style={font=\footnotesize, draw=gray!50},
    grid=major,
    grid style={dotted, gray!40},
    tick label style={font=\footnotesize},
    label style={font=\small},
  ]
  \addplot[thick, blue!70!black, smooth] coordinates {
    (0,45)(1000,52)(2000,58)(3000,63)(4000,70)
    (5000,73)(6000,75)(7000,77)(8000,78)(10000,78)
  };
  \addlegendentry{$R_{\mathrm{cor}}{>}0$ fraction}
  \addplot[thick, red!70!black, dashed, smooth] coordinates {
    (0,8)(500,6.2)(1000,4.5)(2000,3.5)(3000,3.1)
    (4000,2.9)(5000,2.8)(6000,2.7)(8000,2.55)(10000,2.4)
  };
  \addlegendentry{$R_{\mathrm{cpl}}$ trigger rate}
  \end{axis}
  \end{tikzpicture}
  \caption{Reward signal dynamics during GRPO training.
    The fraction of group-8 samples with $R_{\mathrm{cor}}{>}0$ (solid)
    rises from 45\% at SFT initialization to 78\% at convergence,
    confirming a non-degenerate reward signal throughout training.
    The $R_{\mathrm{cpl}}$ trigger rate (dashed) falls below 3\%
    after 1{,}000 steps; residual grey-area violations motivate the
    subsequent DPO stage.}
  \label{fig:dynamics}
\end{figure}
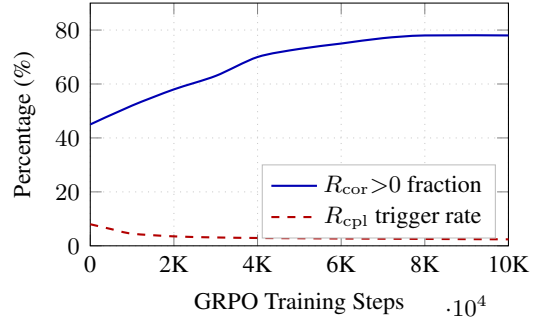

Figure~\ref{fig:dynamics} plots the reward signal dynamics throughout
GRPO training. The fraction of group-8 samples with $R_{\text{cor}}{>}0$
rises from 45\% (SFT initialization) to 78\% by convergence,
confirming the reward signal remains non-degenerate throughout.
$R_{\text{cpl}}$ triggers on $<$3\% of trajectories after 1,000 steps;
the residual grey-area violations motivate the subsequent DPO stage.

\subsection{Model Size and Training Efficiency}

Table~\ref{tab:size} reports the accuracy--latency trade-off
across three Qwen3 model sizes, all trained with the full
ToolRLA pipeline.

\begin{table}[h]
\centering
\small
\begin{tabular}{lccc}
\toprule
\textbf{Model} & \textbf{TCR} $\uparrow$ & \textbf{TIER} $\downarrow$
  & \textbf{Latency} $\downarrow$ \\
\midrule
Qwen3-8B   & 84.0 & 20.5 & 1.1s \\
Qwen3-14B  & 91.0 & 14.0 & 1.6s \\
Qwen3-32B  & 93.5 & 12.2 & 3.8s \\
\bottomrule
\end{tabular}
\caption{Model size ablation on FA-Bench (all trained with ToolRLA).}
\label{tab:size}
\end{table}

Qwen3-14B closes 84\% of the TCR gap between 8B and 32B while
maintaining sub-2s latency. The 32B variant provides only marginal
gains (+2.5pp TCR, $-$1.8pp TIER) at 2.4$\times$ higher inference
cost, making it impractical under our latency budget.
GRPO convergence stabilizes at $\approx$8,000 steps (78\% of samples
with positive advantage), faster than typical PPO schedules
requiring $>$50K rollouts, attributable to the fine-grained reward
reducing sparsity by providing dense per-dimension gradient signal
even for partially correct trajectories.

\section{Discussion and Limitations}

\paragraph{Generalizability.}
Public benchmark results on ToolBench and API-Bank suggest the
multiplicative reward structure transfers beyond the financial domain.
The prerequisite-chain insight applies to any setting where
tool selection errors and parameter errors have qualitatively
different semantics and domain constraints require explicit priority ordering.

\paragraph{Limitations.}
\textit{Sandbox fidelity}: our reward depends on a weekly-synchronized
data replica; a backend API field rename once caused zero accuracy scores
for two days, underscoring the need for automated schema consistency checks.
\textit{FA-Bench privacy}: the internal benchmark cannot be released;
reproducibility relies on public benchmark results.
\textit{Annotation cost}: the DPO compliance dataset required $\sim$3 weeks
of part-time expert annotation; inter-annotator agreement was 84\%
before arbitration, reflecting genuine boundary ambiguity.
\textit{Modality}: the current system handles text-only inputs;
multimodal extensions (chart images, scanned documents) would require
additional reward signals.

\paragraph{Future Directions.}
Promising extensions include multimodal tool integration,
event-triggered proactive advisory (non-episodic RL),
and lightweight per-advisor personalization via LoRA fine-tuning.

\section{Conclusion}

We presented ToolRLA, a three-stage post-training framework
for tool-integrated agents in domain-specific settings.
The central contribution is a fine-grained multiplicative reward function
that evaluates tool invocation quality along four dimensions and encodes
task-specific priority orderings as inductive biases in the reward landscape.
Ablation studies demonstrate that multiplicative composition
of the correctness reward accounts for 7 percentage points of
TIER improvement over additive alternatives, and that the
three-stage pipeline (SFT $\to$ GRPO $\to$ DPO) is strictly
better than any prefix thereof.
Deployed on a production financial advisory copilot over three
months, ToolRLA delivers a 47\% improvement in task completion
rate, a 63\% reduction in tool invocation errors, and a 93\%
reduction in regulatory violations.
These results establish structured, semantics-aware reward decomposition
as a practically effective direction for tool-integrated reinforcement
learning beyond binary feedback signals.

\bibliography{references}

\appendix

\section{Hallucination Defense: Implementation Details}
\label{app:hallucination}

We employ a three-layer defense against hallucinated tool invocations.

\paragraph{Layer 1: Prompt-level tool enumeration.}
The system prompt enumerates all valid tool names and their JSON Schema
definitions at every inference step.  This gives the model a grounded
vocabulary of admissible actions and reduces out-of-vocabulary tool
generation at the output-distribution level.

\paragraph{Layer 2: Runtime tool-name validation.}
Before dispatching any action to a backend API, the execution engine
checks the generated \texttt{tool\_name} against the registered tool
registry.  An unrecognized name returns a structured error observation,
e.g., \texttt{\{"error": "unknown\_tool", "valid\_tools": [...]\}},
which the model can read and self-correct within the same trajectory.

\paragraph{Layer 3: Error-recovery demonstrations in the SFT corpus.}
Approximately 5\% of SFT trajectories explicitly demonstrate the
recover-from-hallucination pattern: the model emits an invalid tool
name, receives the error observation, and then selects the correct tool.
This teaches the model that hallucination is recoverable rather than
terminal, improving robustness under distribution shift.

\paragraph{Effect.}
Combined, these three layers reduce hallucinated tool invocations from
$\sim$8\% (SFT initialization) to $<$1\% after GRPO training, as
measured on the FA-Bench held-out set.

\end{document}